\documentclass[sigconf]{acmart}

\copyrightyear{2023}
\acmYear{2023}
\setcopyright{acmlicensed}
\acmConference[MM '23] {Proceedings of the 31st ACM International Conference on Multimedia}{October 29--November 3, 2023}{Ottawa, ON, Canada.}
\acmBooktitle{Proceedings of the 31st ACM International Conference on Multimedia (MM '23), October 29--November 3, 2023, Ottawa, ON, Canada}
\acmPrice{15.00}
\acmISBN{979-8-4007-0108-5/23/10}
\acmDOI{10.1145/3581783.3612052}

\begin{document}

\title{Parsing is All You Need for Accurate Gait Recognition in the Wild}


\author{Jinkai Zheng}
\authornote{This work was done when Jinkai Zheng was an intern at JD Explore Academy.}
\authornote{Equal contribution.}
\email{zhengjinkai3@hdu.edu.cn}
\affiliation{%
  \institution{Hangzhou Dianzi University}
  \city{Hangzhou}
  \country{China}
}

\author{Xinchen Liu}
\authornotemark[2]
\email{xinchenliu@bupt.cn}
\affiliation{%
  \institution{JD Explore Academy}
  \city{Beijing}
  \country{China}
}

\author{Shuai Wang}
\email{shuaiwang.tai@gmail.com}
\affiliation{%
  \institution{Hangzhou Dianzi University}
  \city{Hangzhou}
  \country{China}
}
\affiliation{%
  \institution{Lishui Institute of Hangzhou Dianzi University}
  \city{Lishui}
  \country{China}
}

\author{Lihao Wang}
\email{wanglihao8@hdu..edu.cn}
\affiliation{%
  \institution{Hangzhou Dianzi University}
  \city{Hangzhou}
  \country{China}
}

\author{Chenggang Yan}
\email{cgyan@hdu.edu.cn}
\affiliation{%
  \institution{Hangzhou Dianzi University}
  \city{Hangzhou}
  \country{China}
}

\author{Wu Liu}
\authornote{Corresponding author.}
\email{liuwu@live.cn}
\affiliation{%
  \institution{JD Explore Academy}
  \city{Beijing}
  \country{China}
}


\renewcommand{\shortauthors}{Jinkai Zheng et al.}
\begin{abstract}
Binary silhouettes and keypoint-based skeletons have dominated human gait recognition studies for decades since they are easy to extract from video frames.
Despite their success in gait recognition for in-the-lab environments, they usually fail in real-world scenarios due to their low information entropy for gait representations.
To achieve accurate gait recognition in the wild, this paper presents a novel gait representation, named \textbf{Gait Parsing Sequence (GPS)}.
GPSs are sequences of fine-grained human segmentation, i.e., human parsing, extracted from video frames, so they have much higher information entropy to encode the shapes and dynamics of fine-grained human parts during walking.
Moreover, to effectively explore the capability of the GPS representation, we propose a novel human parsing-based gait recognition framework, named \textbf{ParsingGait}.
ParsingGait contains a Convolutional Neural Network (CNN)-based backbone and two light-weighted heads.
The first head extracts global semantic features from GPSs, while the other one learns mutual information of part-level features through Graph Convolutional Networks to model the detailed dynamics of human walking.
Furthermore, due to the lack of suitable datasets, we build the first parsing-based dataset for gait recognition in the wild, named \textbf{Gait3D-Parsing}, by extending the large-scale and challenging Gait3D dataset.
Based on Gait3D-Parsing, we comprehensively evaluate our method and existing gait recognition methods. 
Specifically, ParsingGait achieves a $17.5\%$ Rank-1 increase compared with the state-of-the-art silhouette-based method.
In addition, by replacing silhouettes with GPSs, current gait recognition methods achieve about $12.5\% \sim 19.2\%$ improvements in Rank-1 accuracy.
The experimental results show a significant improvement in accuracy brought by the GPS representation and the superiority of ParsingGait. 
The code and dataset are available at \url{https://gait3d.github.io/gait3d-parsing-hp}.
\end{abstract}

%


\maketitle



\begin{figure}[t]
  \centering
   \includegraphics[width=0.95\linewidth]{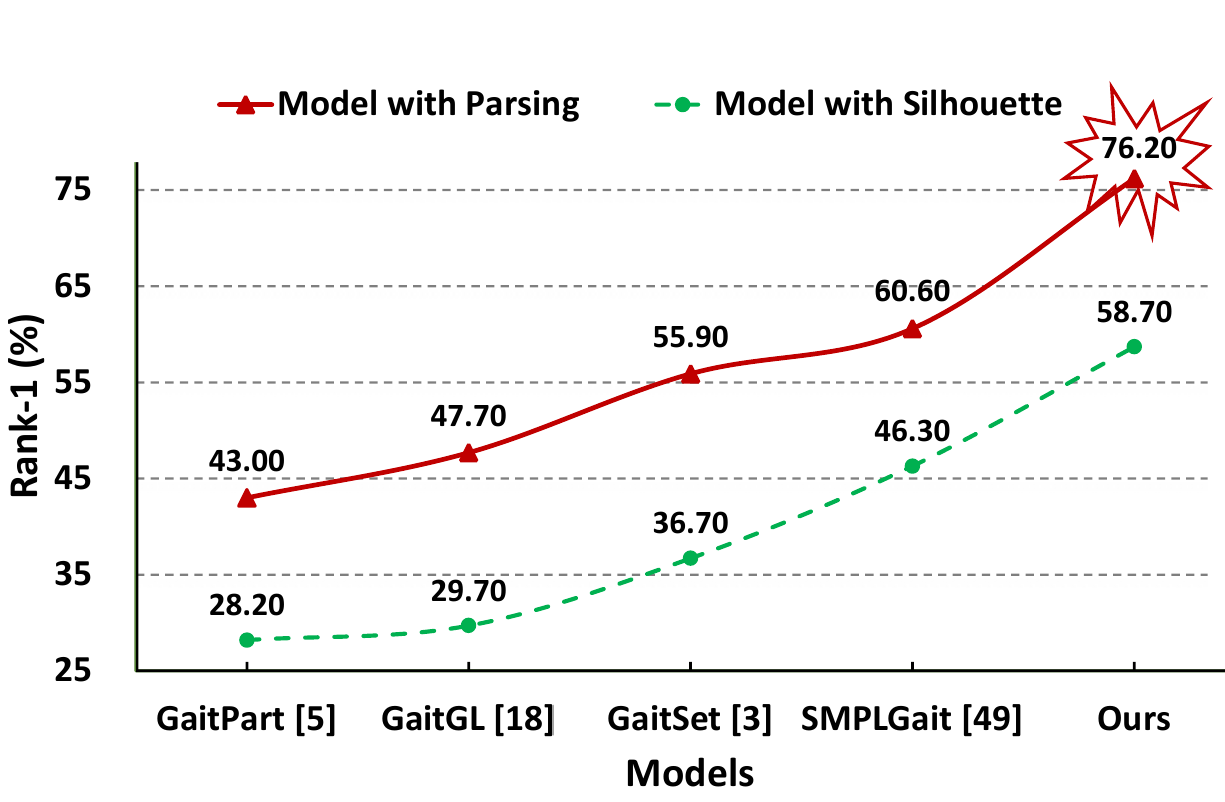}
   \caption{By exploiting parsing in place of silhouettes, contemporary gait recognition methods have attained a substantial improvement of $12.5\% \sim 19.2\%$ in Rank-1 accuracy. Furthermore, our ParsingGait achieves a notable $15.6\%$ enhancement in performance compared to other state-of-the-art methods using parsing as input. (Best viewed in color.)
   }
   \label{fig:figure1} \vspace{-5mm}
\end{figure}

\section{Introduction}
Gait recognition aims to identify the target person by the motion pattern of human walking. 
Compared with face and fingerprint, gait has the advantages of being remotely accessible, non-contacting, and hard to disguise, which makes it a promising biometric for applications like ID authentication and social security~\cite{cvpr/NiyogiA94, csur/WanWP19}. 

Existing gait representations for gait recognition are dominated by silhouette sequences~\cite{icpr/YuTT06}, Gait Energy Images (GEIs)~\cite{pami/HanB06}, 2D skeletons~\cite{icip21_gaitgraph}, 3D skeletons~\cite{pr/LiaoYAH20}, etc. 
Although these representations perform well in laboratory environments, they usually fail in unconstrained situations, i.e., gait recognition in the wild~\cite{Zhu_2021_ICCV, cvpr/gait3d_v1}. 
This is because the information entropy of binary silhouettes or 2D/3D keypoints is too low to encode effective shapes and dynamics of the human body during walking.
Therefore, it is necessary to explore a new representation with higher information entropy for gait recognition in the wild. 

In addition, existing gait recognition methods can be divided into two categories: model-based and model-free approaches. 
The model-based approaches often take 2D/3D skeletons as gait representations. 
However, due to the loss of much useful gait information, these methods are often inferior to model-free approaches in performance~\cite{Zhu_2021_ICCV, cvpr/gait3d_v1}. 
The model-free methods, i.e., the appearance-based methods, mainly take silhouette sequences or GEIs as inputs. 
In early years, GEINet~\cite{icb/ShiragaMMEY16} obtained the Gait Energy Images (GEIs) by compressing the gait sequences in the temporal dimension and then trained the Convolutional Neural Network (CNN) in a classification manner. 
Compared with GEIs, silhouette sequences have higher information entropy to model the appearance, body shape, and belongings of people.
Therefore, the gait recognition community has been dominated by silhouettes in recent years. 
For example, 
GaitSet~\cite{aaai/ChaoHZF19} first regarded a silhouette sequence as a set and directly utilized CNNs to extract the spatial-temporal features from it. 
GaitPart~\cite{cvpr/FanPC0HCHLH20} proposed a novel network to horizontally split the gait silhouettes into strips and extract detailed features from each strip. 
GaitGL~\cite{Lin_2021_ICCV} employed 3D convolution to extract global and local features from the gait silhouette sequence. 
Despite their success on widely adopted in-the-lab gait recognition datasets like CASIA-B~\cite{icpr/YuTT06} and OU-MVLP~\cite{ipsjtcva/TakemuraMMEY18}, they can not work well on recent in-the-wild datasets like GREW~\cite{Zhu_2021_ICCV} and Gait3D~\cite{cvpr/gait3d_v1}. 
This is also caused by the low information entropy of the binary silhouette which only contains a global profile of the whole body without any details of the fine-grained body parts.
Therefore, we aim to model the shapes and dynamics of fine-grained body parts during walking with the part-level gait representation.

To overcome the low information entropy of existing gait representations, we propose a novel gait representation, named Gait Parsing Sequence (GPS), which can be easy to extract by the off-the-shelf human parsing models~\cite{hrnet_segmentation, cdgnet}.
Compared with current representations for gait recognition, the advantages of the GPS are two-fold:
1) the GPS contains higher information entropy at the pixel level which can encode more information about the walking of people. 
2) the GPS represents dense part-level knowledge that enables the model to learn fine-grained dynamics of various parts of the human body during walking. 

To effectively explore the capability of the GPS, we propose a novel human parsing-based gait recognition framework, named ParsingGait, which takes only GPSs as the input instead of the silhouettes or skeletons. 
ParsingGait contains a Convolutional Neural Network (CNN)-based backbone and two light-weighted heads. 
The CNN-based backbone adopts a ResNet-like structure to effectively learn the mid-level features from the GPSs while alleviating gradient vanishing.
The first head of ParsingGait is designed to extract global semantic features of gait, including clothing, body shape, etc., from the GPSs.
To discover the detailed part-level dynamics, we propose the important cross-part modeling head.
The cross-part modeling head not only can learn detailed features of individual body parts but also can learn the mutual information among different parts through a multi-layer Graph Convolutional Network (GCN).
By this means, the part-level features can be more robust to real-world challenges like occlusions, background clutter, etc.
Finally, by integrating the features from the two heads, a discriminative gait representation can be learned for accurate gait recognition in the wild. 

Due to the lack of suitable datasets, we also build the first human parsing-based dataset for gait recognition in the wild, named Gait3D-Parsing. 
The Gait3D-Parsing dataset is an extension of the large-scale and challenging Gait3D dataset which is collected from an in-the-wild environment~\cite{cvpr/gait3d_v1}.
To obtain high-quality parsing results for Gait3D, we randomly select several thousand images from Gait3D and manually annotate the fine-grained parts to train a human parsing network. 
With this trained human parsing network, we extract all human parsing results for the images in Gait3D to obtain the GPSs of the gait sequences in Gait3D. 
Based on Gait3D-Parsing, we conduct comprehensive benchmarks to evaluate our method and existing gait recognition methods. 
In particular, ParsingGait achieves a 17.5\% Rank-1 increase compared with the SOTA method. 
Moreover, all model-free gait recognition methods achieve $12.5\% \sim 19.21\%$ improvements in Rank-1 accuracy by replacing the silhouette with the GPS, which shows the significant improvement of the GPS for gait recognition in the wild. 

In summary, the contributions of this paper are as follows:
\begin{itemize}
    \item To the best of our knowledge, we are the first work to explore a new simple but effective gait representation for gait recognition, i.e., Gait Parsing Sequence, which contains higher information entropy than existing silhouette or skeleton-based representations. 
    \item We propose a novel parsing-based gait recognition framework, named ParsingGait, which not only can learn discriminative features from the fine-grained body parts but also can model the relations among different parts, which makes the features more robust in real-world scenes.
    \item We build the first large-scale parsing-based gait dataset, named Gait3D-Parsing, which can greatly facilitate the development of gait recognition research and applications. 
\end{itemize}

\section{related work}
\label{sec:related}
In this section, we first introduce the existing gait representations. 
Then, the gait recognition methods are briefly surveyed. 
Finally, we summarize the human parsing methods. 

\subsection{Gait Representations}
The existing methods for gait representations are primarily dominated by gait silhouette sequences, Gait Energy Image (GEI), 2D/3D skeleton, and 3D SMPL\&Mesh.
These representations form the basis of current publicly available gait recognition datasets, which can be divided into two main series: the CASIA series~\cite{pami/WangTNH03, icpr/YuTT06, icpr/TanHYT06} and the OU-ISIR series~\cite{cvpr/TsujiMY10, pr/HossainMWY10, accv/MakiharaMY10, tifs/IwamaOMY12, ipsjtcva/XuMOLYL17, ipsjtcva/UddinTMTLMY18, tbbis/AnYMWXYLY20}.
The CASIA series was developed in the early stages of gait recognition and facilitated the initial exploration of RGB images and silhouettes for gait representations~\cite{icpr/TanHYT06, pami/HanB06}.
The OU-ISIR series was first built ten years ago and includes various variants such as walking at different speeds~\cite{cvpr/TsujiMY10}, different clothing styles ~\cite{pr/HossainMWY10}, carrying bags~\cite{ipsjtcva/UddinTMTLMY18}, subjects of different ages~\cite{ipsjtcva/XuMOLYL17}, and annotations of 2D pose~\cite{tbbis/AnYMWXYLY20}.
The OU series provides gait representations of gait silhouette sequence, GEI, 2D/3D skeleton, and 3D SMPL simultaneously. 
Due to their large population, the OU-LP~\cite{tifs/IwamaOMY12} and OU-MVLP~\cite{ipsjtcva/TakemuraMMEY18} also became the most popular datasets for current research.

Recent studies have shown that existing gait recognition datasets were collected in restricted environments such as laboratories ~\cite{icpr/YuTT06, tifs/IwamaOMY12} or small designated areas on campuses~\cite{pami/WangTNH03, pami/SarkarPLVGB05}.
Most recently, researchers started to narrow the gap between in-the-lab research and real-world application.
For example, the GREW~\cite{Zhu_2021_ICCV} and Gait3D~\cite{cvpr/gait3d_v1} are constructed from natural videos collected in unconstrained environments, i.e., in the wild scenes. 
Although existing gait recognition methods perform well on in-lab datasets~\cite{icpr/TanHYT06, tifs/IwamaOMY12, ipsjtcva/TakemuraMMEY18}, they exhibit poor performance on in-the-wild datasets~\cite{Zhu_2021_ICCV, cvpr/gait3d_v1}. 
This is attributed to the low information entropy of existing gait representations, which are dominated by gait silhouettes and are insufficient in representing gait patterns. 
To overcome challenges in real-world scenes, such as complex backgrounds, dynamic routes, and irregular speeds, a new and effective gait representation with high information entropy is necessary for gait recognition in the wild.


\subsection{Gait Recognition Methods}
Gait recognition methods can be classified into two main categories: model-based and model-free approaches~\cite{csur/WanWP19}.
Early methods predominantly utilized model-based approaches, which involved 2D/3D skeletons or a structural human body model~\cite{pr/YamNC04, icb/AriyantoN11}.
For example, Urtasun and Fua~\cite{fgr/UrtasunF04} proposed a gait analysis method that depended on articulated skeletons and 3D temporal motion models.
Zhao~\textit{et al.}~\cite{fgr/ZhaoLLP06} used a local optimization algorithm to track motion for gait recognition.
Yamauchi~\textit{et al.}~\cite{cvpr/YamauchiBS09} developed the first method for walking human recognition using 3D pose estimated from RGB frames.
However, model-based approaches are less effective than model-free methods because they lack valuable gait information such as shape and appearance.

Model-free methods, particularly appearance-based approaches, utilize the gait silhouette sequence or Gait Energy Image (GEI) as input~\cite{cvpr/ZhangT0A0WW19, cvpr/ZhangWL21}. 
Recently, due to the success of deep learning for computer vision tasks~\cite{pami/gongbao_ycg, tomm/lizhisheng_ycg, liu2021recent, csvt/ycg_2022, tomm/tengtong_ycg, mm_LiuLZY020_zheng, tomm/ycg_age_2022}, deep learning-based methods also dominated the performance of gait recognition. 
Notably,  Han~\textit{et al.} proposed consolidating a sequence of silhouettes into a compact Gait Energy Image (GEI)~\cite{pami/HanB06} which subsequent methods widely adopted~\cite{icb/ShiragaMMEY16, pami/WuHWWT17}.
Shiraga~\textit{et al.}~\cite{icb/ShiragaMMEY16} and Wu~\textit{et al.}~\cite{pami/WuHWWT17} proposed to learn effective features from GEIs and significantly outperformed previous methods.
The most recent methods started to learn discriminative features directly from the gait silhouette sequences using larger CNNs or multi-scale structures and achieved state-of-the-art results~\cite{aaai/ChaoHZF19, cvpr/FanPC0HCHLH20, Huang_2021_ICCV, Lin_2021_ICCV}. 
Despite the excellent performance on in-the-lab datasets, these methods usually fail in the wild~\cite{Zhu_2021_ICCV, cvpr/gait3d_v1}. 
This is because existing methods are based on silhouette as input, and gait silhouette sequence is low information entropy, which can not well represent rich gait information. 
The Gait Parsing Sequence (GPS) can not only provide body shape and appearance but also contain the motion of various parts of the human body, which is more robust and discriminative for gait recognition. 
Therefore, this paper aims to explore a parsing-based method for gait recognition in the wild. 

\subsection{Human Parsing Approaches}
Human parsing is a specific task of semantic segmentation for pixel-level classification of human parts~\cite{acmmm_parsing_1, acmmm_parsing_2, acmmm_parsing_3}. 
In recent years, researchers have developed this subfield by exploring the unique human body structures and clothing characteristics. 
For example, 
Wang~\textit{et al.}~\cite{wang2020hierarchical} proposed to combine three kinds of part relations and employed spatial information preservation for efficient and complete human parsing. 
Liu~\textit{et al.}~\cite{cdgnet} designed a novel CDGNet framework that constructs class distribution labels in the horizontal and vertical dimensions to achieve accurate human parsing. 
These methods achieve good performance on existing human parsing datasets such as the Look Into Person (LIP) dataset~\cite{cvpr_GongLZSL17} and the Crowd Instance-level Human Parsing (CIHP) dataset~\cite{eccv_GongLLCYL18}. 
However, the classification of human body parts and clothing in these datasets cannot meet our needs for gait recognition in the wild. 
Therefore, we construct a special fine-grained human parsing dataset, which can be used for gait recognition tasks. 

\section{Gait Parsing Sequence}
In this section, we introduce the definition of the novel representation, i.e., Gait Parsing Sequence (GPS), for gait recognition.
Given a human walking sequence of RGB frames $S=\{\mathbf{I}^i\}^N_{i=1}$, where $\mathbf{I}^i \in \mathbb{R}^{3 \times H \times W}$ is one RGB frame and $N$ is the length of the sequence, we can feed each frame $\mathbf{I}^i$ into a well-trained human parsing network $P(\cdot)$ to obtain the parsing frame $\textbf{x}^i=P(\textbf{I}^i)$.
The human parsing network $P(\cdot)$ can perform the pixel-level classification of predefined $K$ body parts (including the background) for $\textbf{I}^i$.
Therefore, the GPS can be formulated as:
\begin{equation}
\label{equ_gps_definition}
X= \{ \mathbf{x}^i\}^{N}_{i=1},
\end{equation}
where $\textbf{x}^{i}\in \mathbb{R}^{H \times W}$ is the parsing frame.
For each pixel $p^j$ of $\mathbf{x}^i$, we have $p^j \in \{0, 1, ..., K-1\}$ where $0$ denotes the background.

Based on the above definition, we compare the proposed GPS representation with the binary silhouette from the viewpoint of information theory.
Inspired by the definition of information entropy proposed by Claude E. Shannon~\cite{DBLP:journals/bstj/Shannon48}, the pixel-level information entropy of the GPS can be denoted as
\begin{equation}
\label{equ_information_entropy_for_gait_representation}
\mathcal{H} = - \sum_{k=0}^{K} p_k \cdot log(p_k),
\end{equation}
where $p_k$ is the possibility of one pixel belonging to the $k$-th class.
For widely-used definition of body parts for human parsing~\cite{cvpr_GongLZSL17, eccv_GongLLCYL18}, the $K$ is usually larger than ten or more.
So, taking $K=16$ as an example, the information entropy of the GPS can be four bits.
However, for the binary silhouette, each pixel only belongs to the foreground or the background, i.e., the $K=2$, so the information of the binary pixel is only one bit.
Therefore, the pixel-level information entropy of GPS is four times of traditional silhouette-based representation.
Although more classes of parts may cause more classification errors, current SOTA human parsing methods~\cite{cdgnet, hrnet_segmentation} have achieved excellent accuracy on challenging benchmarks, which can guarantee the high quality of parsing results.
Therefore, we consider GPS as a gait representation with higher information entropy to encode the fine-grained dynamics of body parts during walking.
Next, we will present how to effectively exploit the rich information from the GPS representation for gait recognition in the wild. 

\begin{figure*}[t]
  \centering
   \includegraphics[width=0.95\linewidth]{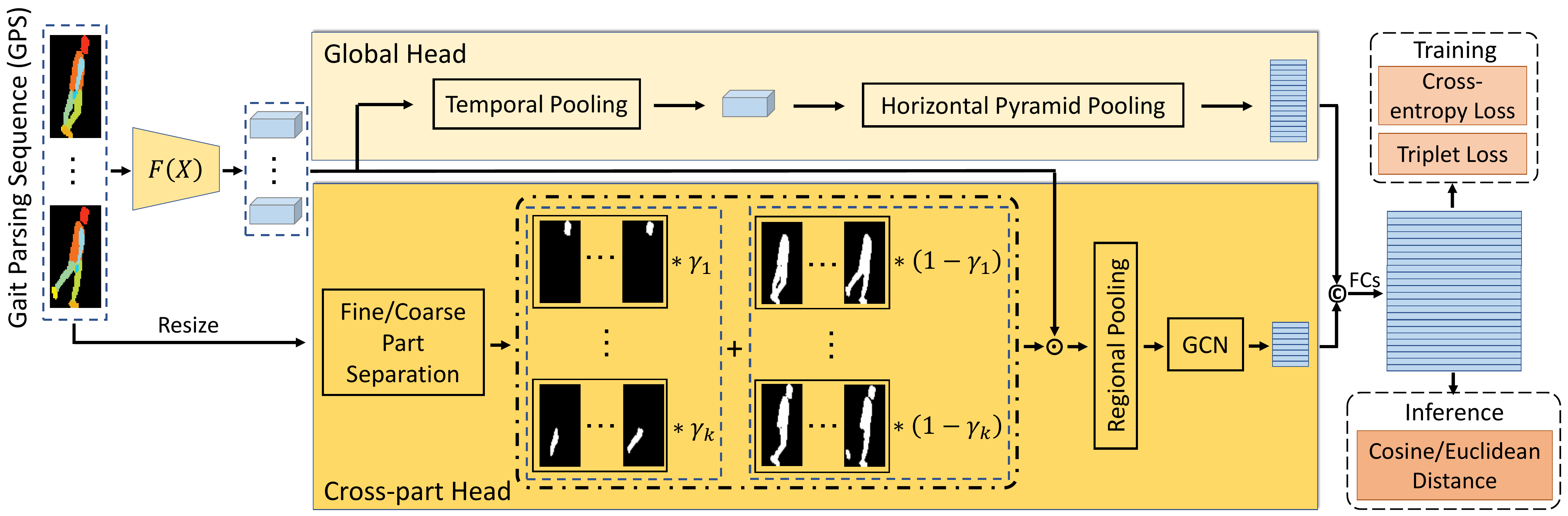}
   \caption{
   The architecture of the ParsingGait framework contains a backbone network and two light-weighted heads. 
   The GPSs are fed into the backbone to obtain the mid-level feature maps.
   The global head is designed to extract global semantic features from the GPSs. 
   The cross-part head not only extracts the discriminative part-level features from individual body parts but also can the mutual information among different parts through a Graph Convolutional Network (GCN). 
   By integrating the features from the two heads, the final gait feature can be obtained for training and inference. (Best viewed in color.)
   }
   \label{fig_parsinggait_framework}
\end{figure*}

\section{ParsingGait for Gait Recognition}
\subsection{Overview}
The overall architecture of the proposed ParsingGait is shown in Figure~\ref{fig_parsinggait_framework}. 
First, the sampled Gait Parsing Sequences (GPSs) are fed into a backbone network, e.g., the ResNet~\cite{resnet} or the Vision Transformers~\cite{swin-transformer} to extract mid-level features.
We formulate $X= \{ \mathbf{x}^i\}^{N}_{i=1}$ as the input GPS, where $x^{i}\in \mathbb{R}^{H \times W}$, $N$ is the length of the sampled sequence, $H$ and $W$ are the height and width of the frame. 
This process can be formulated as:
\begin{equation}
\label{equ_overview}
\mathbf{F}^i = F(\mathbf{x}^i),
\end{equation}
where $F(\cdot)$ is the backbone and $\mathbf{F}^i \in \mathbb{R}^{c \times h \times w}$ is the frame-level feature map for frame $\mathbf{x}^i$, $c$, $h$, and $w$ are the number of channels, height, and width of the feature map, respectively. 
In our implementation, we employ the ResNet-9 in GaitBase~\cite{opengait} as the backbone due to its effectiveness.

To learn discriminative part-level features, we design two light-weighted heads. 
For the first head, i.e., the global head, we aim to extract the global features, which can learn the information on appearance and shape from each frame. 
For the second head, i.e., the cross-part head, we model the relationship among different parts. 
Since the spatial relation of the body parts can be naturally represented as a graph, the Graph Convolutional Network (GCN) is adopted to learn the cross-part knowledge.
By this means, the regional features of individual parts can be propagated across different parts.
Finally, the output features of the two heads are aggregated for sequence-to-sequence matching in training or inference. 
Next, we will introduce the two heads in detail. 

\subsection{The Global Head}
In the global head of the ParsingGait, we aim to learn the appearance knowledge of humans like 2D spatial information and global semantic knowledge. 
As is shown in Figure~\ref{fig_parsinggait_framework}, we first utilize temporal pooling to integrate the temporal-level features for computing considerations. 
Next, the Horizontal Pyramid Pooling (HPP) in GaitSet~\cite{aaai/ChaoHZF19} is adopted to generate more refined gait features. 
The HPP operation first horizontally splits features into strips, and then reduces the spatial dimension using max pooling and mean pooling. 
The above operations can be written as: 
\begin{equation}
\label{equ_global_branch}
\mathbf{F}^{i}_{gl} = H(T(\mathbf{F}^{i})),
\end{equation}
where $T(\cdot)$ is the temporal pooling, which adopts max pooling in the temporal dimension. 
$H(\cdot)$ means the HPP operation. 

\subsection{The Cross-part Head}
\label{sectino_cross_part_branch}
The cross-part head is designed to construct the correlation between parsed human part features. 
After the operation in Equation~\ref{equ_overview}, a semantic feature map $\mathbf{F}^i \in \mathbb{R}^{c \times h \times w}$ is obtained. 
We resize the input parsed mask, $M \in N ^{H \times W}$ to $\tilde{M} \in N ^ {h \times w}$ of the same size as the feature map $\mathbf{F}^i$. 
Considering that the proportion of some parts is very small, or even some parts may be occluded, we propose a learnable strategy to extract the regional feature map, which can automatically control the proportion of front and back scenes. 
The regional feature map $\mathbf{F}_{k}^i$ for part $k$ can be extracted by: 
\begin{equation}
\label{equ_get_mask_part_features}
\mathbf{F}_{k}^i = \gamma_k \ast \mathbf{F}^i \odot \tilde{M}_k + (1- \gamma_k) \ast \mathbf{F}^i \odot (1- \tilde{M}_k),
\end{equation}
where $\gamma_k$ is a learnable factor to control the weight of the k-th part feature, $\odot$ is the element-wise multiplication, and $\tilde{M}_k$ is the resized parsed mask of the k-th part. 
Then, we perform Regional Pooling (RP) to obtain a feature vector for each part. 
In particular, RP consists of max-pooling and mean-pooling. 

To better model the correlation between various human body part features, we construct two different scale graph structures, i.e., fine-class graph and coarse-class graph, to build the part-neighboring relationship. 
In particular, the fine-class graph directly treats 11 parts as graph nodes. 
The coarse-class graph is built for more high-order graph modeling. 
Specifically, we reorganized the graph structure, combining the original 11 fine parts into 5 coarse parts according to the topological structure. 
More details are described in detail in Section~\ref{section_implementations_protocols}. 
Both fine-class graph and coarse-class graph can be formulated by an adjacent matrix $A \in \mathbb{R}^{C \times C}$, where $C$ is the number of nodes, i.e., $C=11$ in fine-class graph and $C=5$ in the coarse-class graph in our implementation. 
The adjacent matrix $A(I, j) = 1$ if part $i$ is neighbor to part $j$. 
We adopt the GCN~\cite{gcn} to model the mutual information between different body part features. 
Specifically, we employ a two-layer GCN, and each layer $L$ can be presented as: 
\begin{equation}
\label{qua_gcn}
X^{(L+1)} = \sigma \left (D^{-\frac{1}{2}} A D^{-\frac{1}{2}} X^{(L)} W^{(L)} \right ),
\end{equation}where $X^{(L+1)}$ is the output feature of the L-th layer, $A \in \mathbb{R}^{C \times C}$ is the adjacent matrix, $D \in \mathbb{R}^{C \times C}$ is the degree matrix of $A$, $W^{(L)}$ is the learnable parameters of the layer $L$. 
The initial feature $X^{(0)}$ is obtained by the regional features. 
At last, the output feature $X^{(L+1)}$ is also performed by temporal pooling to obtain the final output feature $\mathbf{F}^{i}_{cp}$. 

\subsection{Training and Inference}
After obtaining $\mathbf{F}^{i}_{gl}$ and $\mathbf{F}^{i}_{cp}$, we first concatenate these two features and then employ several independent Fully Connected layers (FCs) for further feature mapping. 
The final features are used for training and inference. 

In general, our ParsingGait framework is trained in an end-to-end manner.
The network of our framework is optimized by a loss function with two components:
\begin{equation}
\label{equ_loss}
L=\alpha L_{tri}+\beta L_{ce},
\end{equation}
where $L_{tri}$ is the triplet loss, $L_{ce}$ is the cross entropy loss.
$\alpha$ and $\beta$ are the weighting parameters. 

During inference, the cosine/euclidean distance is used to measure the similarity between a query-gallery pair. 

\section{Construction of Gait3D-Parsing}
Due to the lack of a suitable dataset, we build a large-scale human parsing-based dataset for gait recognition in the wild. 
We first introduce the Gait3D dataset and describe why we want to build a parsing-based dataset based on it. 
Next, we construct a Fine-grained Human Parsing (FHP) dataset, which is annotated as a subset from Gait3D to obtain accurate pixel-level parsed human parts. 
Then we carry out the empirical study of human parsing experiment on the FHP dataset. 
Finally, we build a high-quality human parsing-based dataset, named Gait3D-Parsing, from the Gait3D dataset by using the best-trained model of human parsing. 
Next, we will describe the above in detail. 

\subsection{The Gait3D Dataset}
The Gait3D dataset~\cite{cvpr/gait3d_v1} is a newly proposed gait in-the-wild dataset. 
It includes 4,000 subjects, 25,309 sequences, and 3,279,239 frame images in total, which are extracted from 39 cameras in an unconstrained indoor scene, i.e., a large supermarket. 
This dataset also has the following special factors: complex background, 3D viewpoint, irregular walking speed, dynamic route, and occlusion. 
All these factors make the Gait3D a great challenging and representative dataset for gait recognition in the wild. 
Based on the above features, we will extract parsing data from Gait3D, to construct a parsing-based gait dataset. 


\subsection{The FHP Dataset}
To ensure accurate human parsing results on 4,000 subjects in the Gait3D dataset, we randomly sampled one RGB image below each subject of Gait3D. 
We label 11 fine-grained parts, including head, torso, left-arm, right-arm, left-hand, right-hand, left-leg, right-leg, left-foot, right-foot, and dress. 
Then, the annotation is performed manually by annotators. 
During annotation, all labels are checked by two rounds to guarantee high quality. 
Finally, we obtain the Fine-grained Human Parsing (FHP) dataset containing 4,000 images with 11-part masks. 

More details about FHP can be found in \textbf{the supplementary material}. 


\begin{table}[t]
\centering
\caption{Quantitative results on the test set of FHP dataset. } 
\begin{tabular}{l|ccc}
Methods                      			& Pixel Acc & Mean Acc & Mean IoU \\ \midrule[1.5pt]
PSPNet~\cite{pspnet}	                & 93.16 & 78.33 & 67.12 \\
HRNet~\cite{hrnet_segmentation}      	& 93.46 & 78.04 & 67.37 \\
CDGNet~\cite{cdgnet}		            & 94.13 & 82.03 & 71.32 \\
\end{tabular} 
\label{tab_results_human_parsing_on_fhp_dataset}
\end{table}



\subsection{Empirical Study on FHP}
In this section, we evaluate three semantic segmentation algorithms on the FHP dataset, which are PSPNet~\cite{pspnet}, HRNet~\cite{hrnet_segmentation}, and CDGNet~\cite{cdgnet}. 
Specifically, we employ ResNet-101~\cite{resnet} as the backbone for PSPNet and CDGNet. 
We also adopt HRNet-W48 as the backbone of HRNet. 





We unify the input size to (256, 256) for all three methods. 
Moreover, to find the best human parsing model, we split the FHP dataset into train/test sets with 3,600/400 images, respectively. 
We use pixel accuracy (Pixel Acc), mean accuracy (Mean Acc), and mean Intersection-over-Union (mIoU) as the protocol metrics. 

More discussions about the sensitivity to the division of FHP can be found in \textbf{the supplementary material}.

\textbf{Experimental results on FHP. }
The experimental results on the FHP dataset for human parsing are listed in Table~\ref{tab_results_human_parsing_on_fhp_dataset}. 
We can observe that all three methods obtain competitive performance, which gives us the confidence to apply parsing data to the field of gait recognition in the wild. 
In contrast, CDGNet achieves the best performance in all evaluation metrics.
Therefore, we adopt the trained CDGNet to extract the GPSs for the gait sequences of the Gait3D dataset.



\begin{figure*}[ht]
  \centering
   \includegraphics[width=0.95\linewidth]{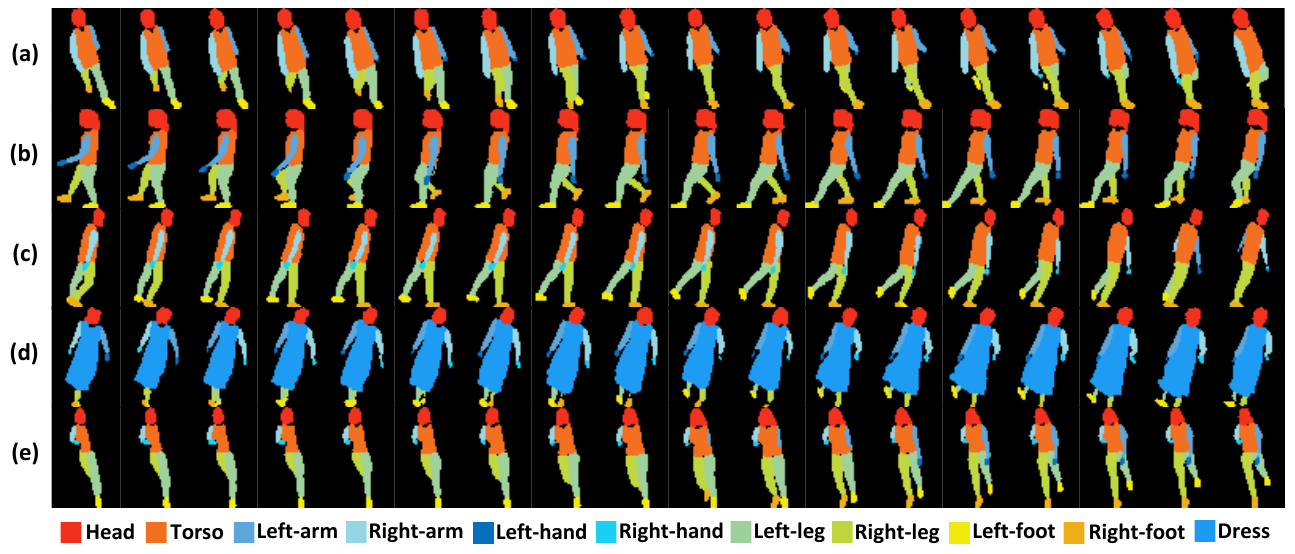}
   \caption{Some visualization of Gait Parsing Sequences (GPSs) in Gait3D-Parsing dataset about (a) child, (b) woman, (c) man, (d) dressing, and (e) occlusion. (Best viewed in color.)
   }
   \label{fig_vis_gait3d_parsing_gps_cut_pkl}
\end{figure*}

\subsection{The Gait3D-Parsing Dataset}
After the above experimental analysis, we select the CDGNet with the best performance as the final human parsing extractor. 
We extract the parsed human parts data from Gait3D~\cite{cvpr/gait3d_v1} dataset to construct the high-quality Gait3D-Parsing dataset. 
The Gait3D-Parsing dataset follows the original subset splitting of Gait3D. 
In particular, the train set has 3,000 IDs, and the test set has 1,000 IDs. 
Meanwhile, 1,000 sequences in the test set are taken as the query set, and the rest of the test set is taken as the gallery set. 
For more details, please refer to the original paper of Gait3D~\cite{cvpr/gait3d_v1}. 

\begin{figure}[t]
  \centering
   \includegraphics[width=1.0\linewidth]{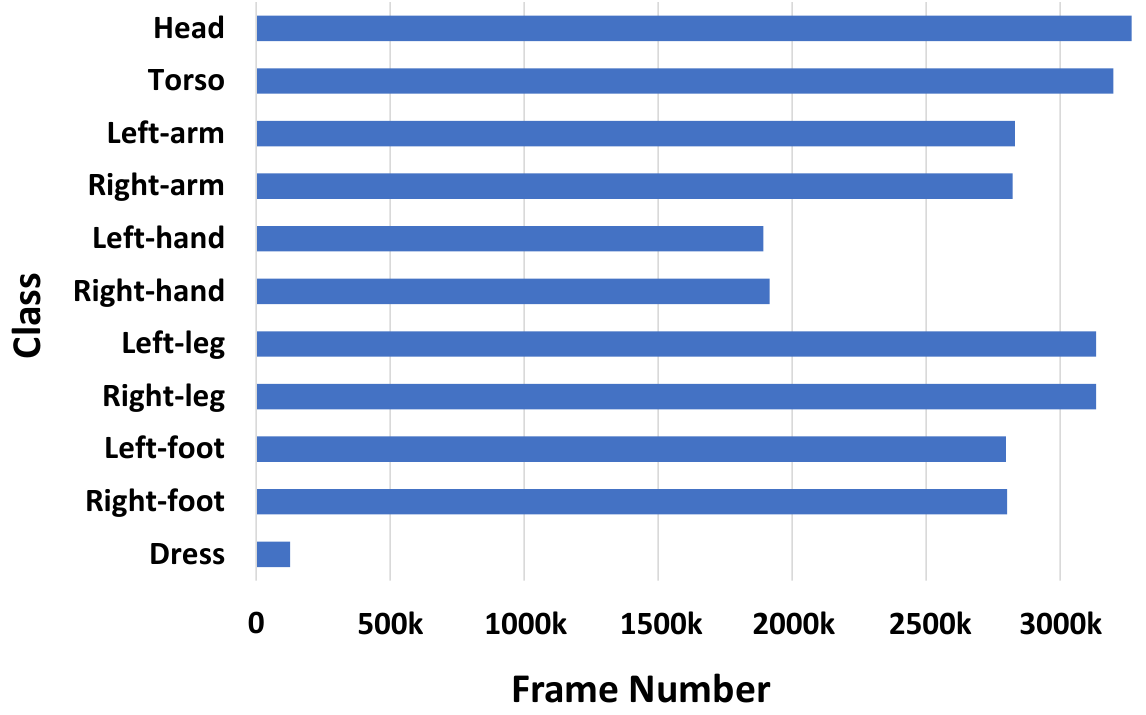}
   \caption{The frame numbers over eleven fine-grained parts on the Gait3D-Parsing dataset. (Best viewed in color.)
   }
   \label{fig_gait3d_parsing_dataset_img_num_over_parts}
\end{figure}

We visualize some Gait Parsing Sequences (GPSs) in the Gait3D-parsing in Figure~\ref{fig_vis_gait3d_parsing_gps_cut_pkl}. 
These examples reflect the high quality of the Gait3D-Parsing dataset,  regardless of gender, age, clothing, occlusion, etc. 
We also illustrate the distribution of frame numbers over 11 fine-grained parts on Gait3D-Parsing, as is shown in Figure~\ref{fig_gait3d_parsing_dataset_img_num_over_parts}. 
We can first observe that the annotations have a relatively balanced distribution over eleven parts, except for the dress class. 
Secondly, we can find the symmetry between left-arm/hand/leg/foot and right-arm/hand/leg/foot, which reveals the characteristic feature of humans walking alternately from left to right. 
Moreover, we can see that the head and torso appear the most, while the dress accounts for the least. 

Finally, it should be pointed out that this is an extension of Gait3D, so we only release the Gait3D-Parsing dataset. 
If readers need additional data about Gait3D, please contact the Gait3D team. 

More statistics of the Gait3D-Parsing dataset can be found in \textbf{the supplementary material}. 

\section{experiments}
In this section, we first introduce the implementation details and evaluation protocols. 
Next, we compare our method with the existing representative gait recognition methods and construct a comprehensive experimental analysis. 
At last, some crucial ablation studies are carried out. 








\subsection{Implementation Details and Protocols}
\label{section_implementations_protocols}
The implementation details and evaluation protocols in our experiments are as follows. 

\textbf{Implementation Details.} 
To verify the effectiveness of our method, some representative gait recognition approaches are introduced for comparison, including GaitSet~\cite{aaai/ChaoHZF19}, GaitPart~\cite{cvpr/FanPC0HCHLH20}, GLN~\cite{eccv/HouCLH20}, SMPLGait~\cite{cvpr/gait3d_v1}, MTSGait~\cite{mtsgait_zheng_mm2022}, GaitBase~\cite{opengait}. 
To make a fair comparison, we follow the training strategy in Gait3D~\cite{cvpr/gait3d_v1} for methods like GaitSet~\cite{aaai/ChaoHZF19}, GaitPart~\cite{cvpr/FanPC0HCHLH20}, GLN~\cite{eccv/HouCLH20}, SMPLGait~\cite{cvpr/gait3d_v1}, and MTSGait~\cite{mtsgait_zheng_mm2022}. 
During training, we train the above models except GLN with the same configuration. 
The batch size is (32, 4, 30), where 32 denotes the number of IDs, 4 denotes the number of training samples per ID, and 30 is the sequence length. 
The models are trained for 1,200 epochs with the initial Learning Rate (LR)=1e-3 and the LR is multiplied by 0.1 at the 200-th and 600-th epochs.
The optimizer is Adam~\cite{corr/KingmaB14} and the weight decay is set to 5e-4.
For GLN, we follow the two-stage training as in~\cite{eccv/HouCLH20}.
The model trained in the first stage is used as the pre-trained model for the second stage.
Both two stages are trained with the same configuration as other methods.
During inference, we use the cosine distance to measure the similarity between each pair of query and gallery sequences. 

For GaitBase~\cite{opengait}, considering the restriction of computing resources, we change the batch size to (32, 2, 30) when reproducing GaitBase~\cite{opengait} and remove augmentations. 
Following the training strategies in GaitBase~\cite{opengait}, we train it for 400 epochs with the initial LR=0.1. 
The LR is multiplied by 0.1 at the 135-th, 270-th, and 335-th epochs. 
The SGD~\cite{sgd_1951} optimizer is employed with the weight decay is 5e-4. 
The Euclidean distance is adopted during inference. 
Because of its excellent performance, we take GaitBase as the baseline model in this paper. 
To make a fair comparison, our proposed method is completely consistent with the training strategy of GaitBase. 
In addition, without a special statement, the input size of all gait recognition models trained in this paper is (64, 44). 

Moreover, the learnable factor $\gamma$ in Equation~\ref{equ_get_mask_part_features} is initialized from 0.75. 
The hyper-parameters in Equation~\ref{equ_loss} are set to $\alpha=1.0$ and $\beta=1.0$. 
The 5 parts of the coarse-class graph mentioned in Section~\ref{sectino_cross_part_branch} are: 
1) head, torso, and dress, 
2) left-hand and left-arm, 
3) right-hand and right-arm, 
4) left-leg and left-foot, 
5) right-leg and right foot. 

\textbf{Protocols.} 
We follow the original evaluation strategies presented in Gait3D~\cite{cvpr/gait3d_v1}. 
The evaluation protocol is based on the open-set instance retrieval setting. 
Given a query sequence, we measure its distance between all sequences in the gallery set.
Then a ranking list of the gallery set is returned by the ascending order of the distance.
We adopt the average Rank-1, Rank-5, and mean Average Precision (mAP) as the evaluation metrics. 

\begin{table}[t]
\centering
\caption{Comparison of the state-of-the-art methods.} 
\begin{tabular}{l|ll}
Methods                      			         & Rank-1 ($\%$)           & mAP ($\%$)   \\ \midrule[1.5pt]
GaitSet~\cite{aaai/ChaoHZF19}		             & 36.70                   &30.01  \\
GaitSet + Parsing		                         & 55.90 (19.2 $\uparrow$) & 46.69 (16.68 $\uparrow$)  \\ \midrule
GaitPart~\cite{cvpr/FanPC0HCHLH20}		         & 28.20                   & 21.58  \\ 
GaitPart + Parsing	                             & 43.00 (14.8 $\uparrow$) & 33.91 (12.33 $\uparrow$)  \\ \midrule
GLN~\cite{eccv/HouCLH20}				         & 31.40                   & 24.74  \\
GLN + Parsing				                     & 45.70 (14.3 $\uparrow$) & 38.56 (13.82 $\uparrow$)  \\ \midrule
GaitGL~\cite{Lin_2021_ICCV}	                     & 29.70                   & 22.29  \\ 
GaitGL + Parsing	                             & 47.70 (18.0 $\uparrow$) & 36.23 (13.94 $\uparrow$)  \\ \midrule
SMPLGait~\cite{cvpr/gait3d_v1}                   & 46.30                   & 37.16  \\ 
SMPLGait + Parsing                               & 60.60 (14.3 $\uparrow$) & 52.29 (15.13 $\uparrow$)  \\ \midrule
MTSGait~\cite{mtsgait_zheng_mm2022}              & 48.70                   & 37.63  \\  
MTSGait + Parsing                                & 61.20 (12.5 $\uparrow$) & 52.81 (15.18 $\uparrow$)  \\ \midrule
GaitBase~\cite{opengait}                         & 58.70                   & 49.54  \\  
GaitBase + Parsing                               & 71.20 (12.5 $\uparrow$) & 64.08 (14.54 $\uparrow$)  \\  \midrule
ParsingGait (Ours)                               & \textbf{76.20}          & \textbf{68.15}  \\  
\end{tabular} 
\label{tab_comparison_sota_methods}
\end{table}

\subsection{Comparison of SOTA Methods}
The Experimental results are listed in Table~\ref{tab_comparison_sota_methods}. 
From the results, we can first observe that our ParsingGait outperforms the existing methods and achieves state-of-the-art performance in all evaluation metrics by a large margin. 
The reason is that more precise local features can be provided by decomposing human body parts, and the correlation between parts can be further explored through our cross-part head. 

Moreover, it is surprising to find that with parsing as input, the performance of all existing methods has been improved by a large margin in all metrics. 
For example, the Rank-1 (mAP) increases by 19.2\% (16.68\%) in GaitSet, 14.8\% (12.33\%) in GaitPart, 18.0\% (13.94\%) in GaitGL, 14.3\% (15.13\%) in SMPLGait, etc. 
This fully demonstrates the importance of the Gait Parsing Sequence (GPS) for gait recognition, which is a high information entropy representation providing appearance, body shape, and semantic information. 
This makes us realize the great potential of Gait Parsing Sequence (GPS) in the field of gait recognition, and GPS is likely to become the mainstream representation of gait recognition in the future.

\subsection{Ablation Study}
In this section, we construct the ablation study of the key components in our ParsingGait method. 
We first show the impact of GCN. 
Then, we conduct experiments to analyze the influence of the 11-fine class graph and the 5-coarse class graph. 
Finally, the learnable factor $\gamma$ in Equation~\ref{equ_get_mask_part_features} is analyzed by experiments. 

\begin{table}[t]
\centering
\caption{Analysis of fine-class, coarse-class, and GCN in ParsingGait.} 
\begin{tabular}{ccc|ccc}
Fine-class & Coarse-class & GCN             & Rank-1 & Rank-5 & mAP \\ \midrule[1.5pt]
$\checkmark$ &              & 	            & 70.30 & 86.10 & 62.84 \\ 
             & $\checkmark$ & 	           & 73.80 & 88.60 & 66.35 \\ 
$\checkmark$ &              & $\checkmark$  & 74.00 & 88.70 & 66.27 \\ 
             & $\checkmark$ & $\checkmark$  & \textbf{76.20} & \textbf{89.10} & \textbf{68.15} \\ 
\end{tabular}
\label{tab_ablation_gcn_fine_coarse_graph}
\end{table}

\begin{figure}[t]
  \centering
   \includegraphics[width=0.9\linewidth]{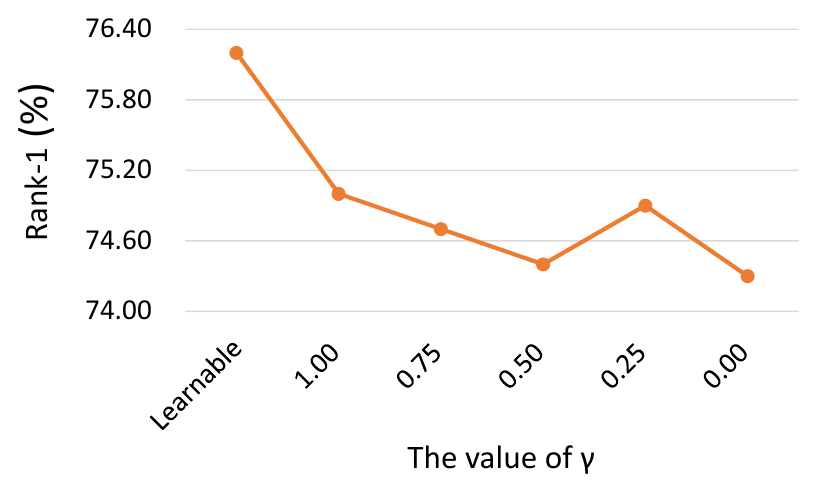}
   \caption{Analysis of different $\gamma$. (Best viewed in color.)}
   \label{fig_ablation_study_gamma}
\end{figure}

\begin{figure*}[t]
  \centering
   \includegraphics[width=0.9\linewidth]{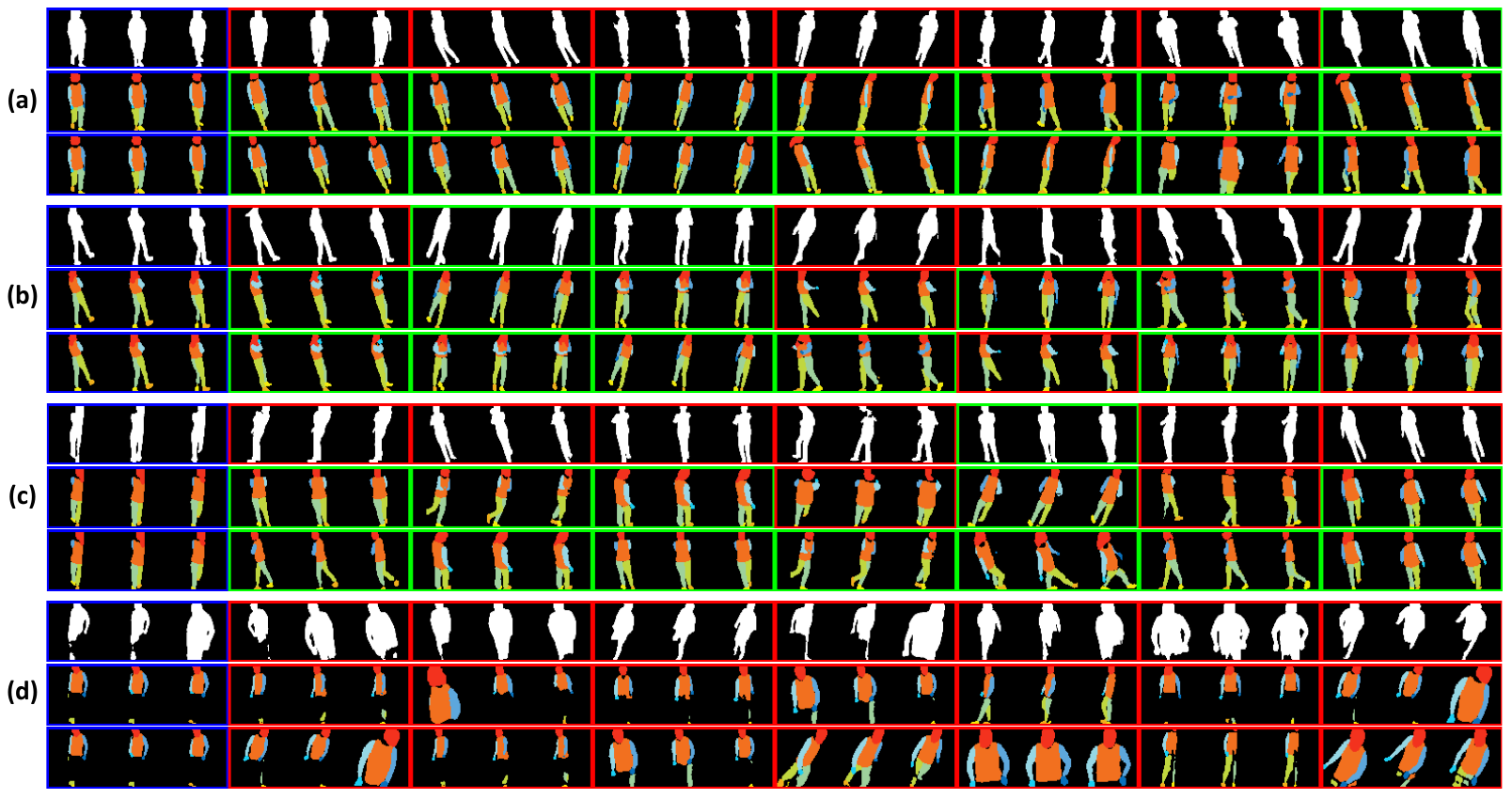}
   \caption{
   Some exemplar results of GaitBase, GaitBase+Parsing, and our ParsingGait. 
   For convenience, we choose the middle frame and the frames with four intervals before and after it for visualization. 
   The blue bounding boxes are queries. 
   The green bounding boxes are the correctly matched results, while the red bounding boxes are the wrong results. 
   The (a) - (d) represent the results under different queries, where the first row of each is the search result of GaitBase, the second row is the result of GaitBase+Parsing, and the third row is the result of ParsingGait.
   (Best viewed in color.)
   }
   \label{fig_vis_exemplar_results}
\end{figure*}

\textbf{The impact of GCN.} 
We try to remove GCN from the cross-part head, directly employ the temporal pooling on regional features, and carry out subsequent operations with the output of the global head. 
The results are listed in Table~\ref{tab_ablation_gcn_fine_coarse_graph}, we can find that GCN plays an important role in both fine-class graph and coarse-class graph structures. 
Specifically, when GCN is added, the Rank-1 accuracy increases by 3.70\%/2.40\% based on fine/coarse-class graph. 

\textbf{Fine-class graph vs. Coarse-class graph.} 
We also compare the performance between fine-class graph and coarse-class graph structures. 
As is shown in Table~\ref{tab_ablation_gcn_fine_coarse_graph}, we can observe that the result of the coarse-class graph is better than that of the fine-class graph. 
The main reason is caused by occlusion. 
Specifically, the probability of each node appearing in each frame in the coarse-class graph is much higher than that in the fine-class graph due to occlusion or viewpoints. 
In addition, we find that the combination of coarse-class graph and GCN obtains the best performance. 

\textbf{The learnable factor $\gamma$.} 
Another important component is the learnable factor $\gamma$. 
We fix other settings and use $0.00 \sim 1.00$ with an increase of 0.25 to analyze the different values for $\gamma$ in Equation~\ref{equ_get_mask_part_features}. 
As illustrated in Figure~\ref{fig_ablation_study_gamma}, we can first find that the performance is best when $\gamma$ is learnable. 
In addition, we observe that the worst performance at $\gamma$ = 0.0 and $\gamma$ = 0.5. 
This indicates that the mask operation in Equation~\ref{equ_get_mask_part_features} needs to retain the target regional features and ensure the difference with other part regions, which is conducive to training the model. 

\subsection{Exemplar Results}
Figure~\ref{fig_vis_exemplar_results} shows several exemplar results of GaitBase~\cite{opengait}, GaitBase+Parsing, and our ParsingGait. 
From Figure~\ref{fig_vis_exemplar_results} (a), we can observe that the model based on the gait silhouette sequence is easy to be confused by the similar body shape.
While the gait parsing sequence, due to its higher information entropy, can satisfy the model to learn more useful gait features so that the model can match more correct results. 
Figure~\ref{fig_vis_exemplar_results} (b) illustrates that the same pose will also interfere with the silhouette-based model while parsing data can help to avoid this problem. 
From Figure~\ref{fig_vis_exemplar_results} (c), we can see that the results returned by the model based on the silhouette are mostly from a similar viewpoint, and the parsing representation can better alleviate this phenomenon. 
Moreover, our ParsingGait can make the model match more correct results from different viewpoints, even the difficult 3D viewpoints. 
Figure~\ref{fig_vis_exemplar_results} (d) shows a bad case where individuals are seriously occluded. 
This indicates that occlusion is one of the main challenges of gait recognition in the wild. 

More detailed exemplar results can be found in \textbf{the supplementary material}.

\section{conclusion}
In this paper, we present a novel representation for gait recognition, named Gait Parsing Sequence (GPS).
Compared with the traditional binary silhouette-based representation, the GPS contains higher information entropy which encodes the shape and dynamic knowledge of fine-grained body parts during walking.
To effectively utilize the GPS, we proposed a parsing-based gait recognition framework, named ParsingGait, which only takes GPS as the input.
The ParsingGait contains a CNN-based backbone and two elaborately-designed heads. 
One head aims to extract global semantic features from GPSs, while the other head learns the correlation among different body parts via a graph-based GCN. 
Moreover, we build the first parsing-based gait recognition dataset, named Gait3D-Parsing, by providing all GPSs for the gait sequences in the Gait3D dataset.
Extensive experiments on the Gait3D-Parsing dataset show significant improvement in current methods with the GPSs as the input and the superior performance of the proposed ParsingGait framework.

\begin{acks}
This work was supported by the National Key Research and Development Program of China under Grant (2020YFB1406604), 
Beijing Nova Program (20220484063), 
National Nature Science Foundation of China (61931008, U21B2024), 
"Pioneer", Zhejiang Provincial Natural Science Foundation of China (LDT23F01011F01)
\end{acks}


\balance
\bibliographystyle{ACM-Reference-Format}
\bibliography{MM2023}

\newpage

\appendix

\section{Appendix: Details about FHP}
\begin{figure}[t]
  \centering
   \includegraphics[width=1.0\linewidth]{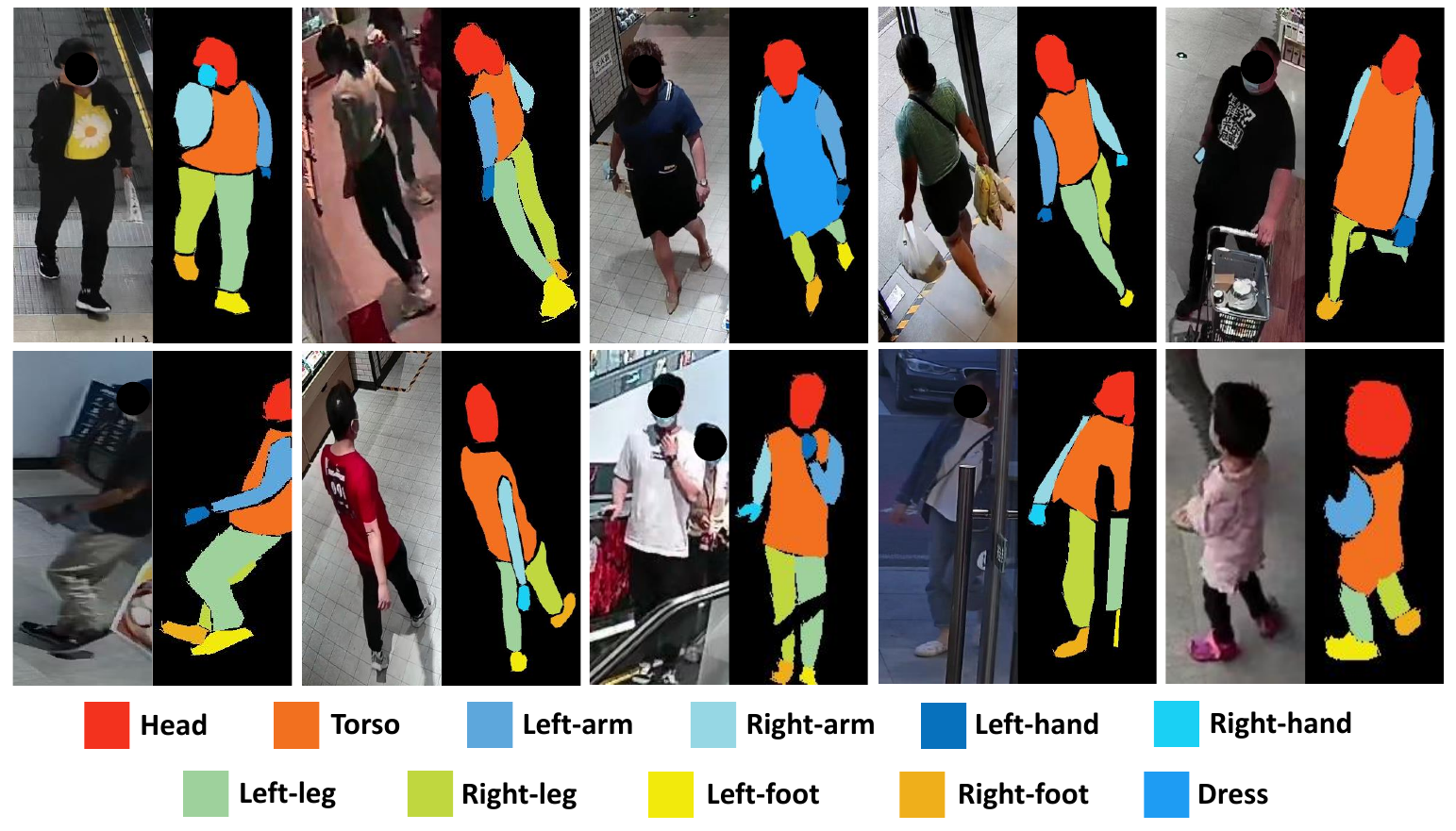}
   \caption{Some images and visualizations of 11 fine-grained annotations on the FHP dataset. (Best viewed in color.)
   }
   \label{fig_vis_fhp_dataset_trainset_anno} \vspace{-3mm}
\end{figure}

While labeling the Fine-grained Human Parsing (FHP) dataset, we do not label belongings such as bags, phones, etc. 
In addition, we only label the most prominent pedestrians in the frame for situations involving multiple people. 
Specifically, we label the pedestrians with the largest proportion in the frame. 
Figure~\ref{fig_vis_fhp_dataset_trainset_anno} illustrates some images and visualizations of 11 fine-grained annotations on the FHP dataset. 

\begin{figure}[t]
  \centering
   \includegraphics[width=1.0\linewidth]{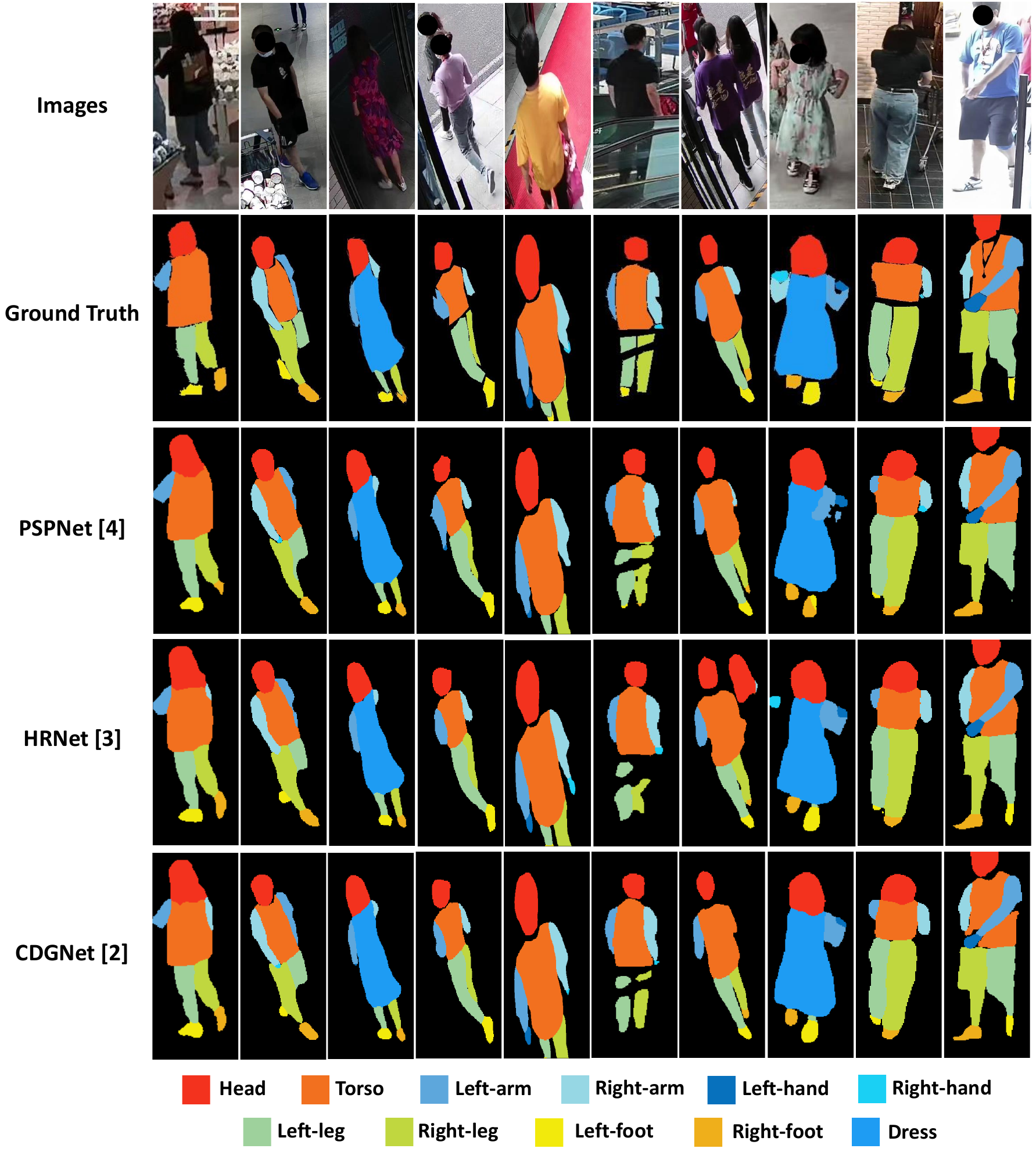}
   \caption{The visualized results on the test set of the FHP dataset. (Best viewed in color.)
   }
   \label{fig_inference_vis_testset_of_fhp_dataset}
\end{figure}

\textbf{Qualitative results. }
To compare the segmentation results of PSPNet~\cite{pspnet}, HRNet~\cite{hrnet_segmentation}, and CDGNet~\cite{cdgnet} more effectively, we visualize the results of all three methods on the test set of the FHP dataset. 
Figure~\ref{fig_inference_vis_testset_of_fhp_dataset} shows that all three methods achieve satisfactory segmentation of human body parts. 
However, in cases involving occlusion, illumination, multiple people, and multiple viewpoints, CDGNet produces the most accurate segmentation results than the other two methods. 

\section{Appendix: Sensitivity to the division of FHP}
To evaluate the sensitivity to the train/test set division of the FHP dataset, we conduct experiments with comprehensive division strategies. 
As we described before, we randomly sample 3,600 images to train the CDGNet, which may contain IDs from the test set of Gait3D. 
To verify whether it will cause information leakage. 
We try using only 3,000 images corresponding to the train set of Gait3D for training CDGNet, and the remaining 1,000 images for testing. 
Considering the decrease in training size, we also add a comparison experiment where 3,000 images are randomly sampled for training the CDGNet and the rest for testing. 
We will name these two division strategies 3k/1k-Gait3D and 3k/1k-Random, respectively. 
Our original division strategy is named 3.6k/0.4k-Random. 

The results for human parsing are listed in Table~\ref{tab_results_human_parsing_with_diff_division}. 
We can observe that the performance of human parsing has a little decrease as the training data drops from 3,600 to 3,000 images. 
In addition, there is almost no difference between different division strategies, when the training data is fixed at 3,000 images. 

We also conduct experiments to evaluate the impact of different division strategies on gait recognition. 
We first utilize the above-trained CDGNet models to extract the different versions of Gait3D-Parsing from Gait3D. 
Then, we train ParsingGait on these different Gait3D-Parsing datasets. 
The experimental results are listed in Table~\ref{tab_impact_diff_divisions_on_parsinggait}. 
We can first find that the performance of gait recognition has a certain decrease as the CDGNet trained from 3,600 images to 3,000 images. 
This shows that the more relevant training data for CDGNet can help improve the performance of gait recognition. 
Moreover, we can observe that the accuracy is stable with `'3k/1k-Gait3D`' and `'3k/1k-Random`' division strategies. 
This demonstrates that even if the training data of CDGNet, i.e., human parsing models, contains the test set of Gait3D, there is no problem of information leakage. 

\begin{table}[t]
\centering
\caption{Quantitative results with different division strategies for training CDGNet on FHP dataset. } 
\begin{tabular}{l|ccc}
Division Strategies     & Pixel Acc & Mean Acc & Mean IoU \\ \midrule[1.5pt]
3k/1k-Gait3D            & 93.63 & 79.97 & 68.57 \\
3k/1k-Random         	& 93.49 & 80.61 & 69.87 \\
3.6k/0.4k-Random      	& 94.13 & 82.03 & 71.32 \\
\end{tabular} 
\label{tab_results_human_parsing_with_diff_division}
\end{table}

\begin{table}[t]
\centering
\caption{The impact of different division strategies on ParsingGait.} 
\begin{tabular}{l|ll}
Division Strategies     & Rank-1 ($\%$)     & mAP ($\%$)   \\ \midrule[1.5pt]
3k/1k-Gait3D		    & 72.40             & 65.28  \\
3k/1k-Random		    & 71.50             & 64.56  \\ 
3.6k/0.4k-Random        & 76.20             & 68.15  \\  
\end{tabular} 
\label{tab_impact_diff_divisions_on_parsinggait}
\end{table}

\section{Appendix: Statistics of Gait3D-Parsing}
\label{more_statistics_of_gait3d_parsing}
\begin{figure}[t]
  \centering
   \includegraphics[width=1.0\linewidth]{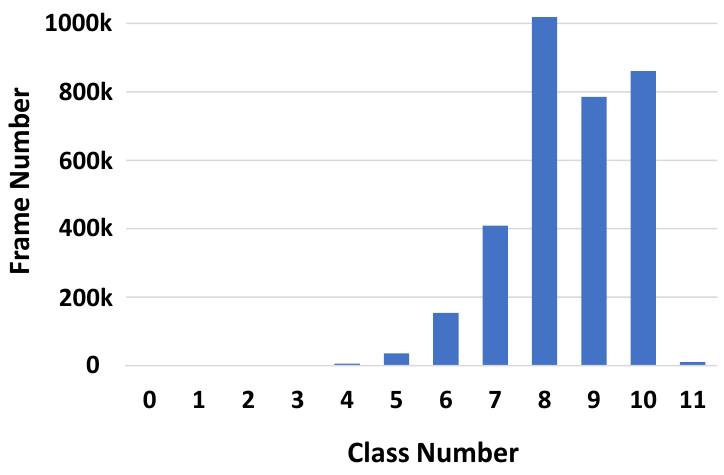}
   \caption{The frame number over the class number on the Gait3D-Parsing dataset. (Best viewed in color.)
   }
   \label{fig_frame_num_over_class_num_of_gait3d_parsing_dataset}
\end{figure}

\begin{figure}[t]
  \centering
   \includegraphics[width=1.0\linewidth]{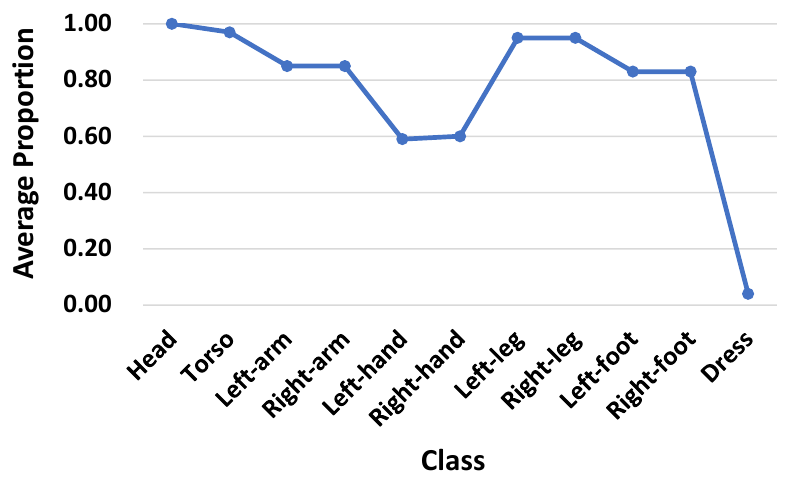}
   \caption{The average proportion of each class within a sequence on the Gait3D-Parsing dataset. (Best viewed in color.)
   }
   \label{fig_mean_proportion_in_one_sequence_over_classes_of_gait3d_parsing_dataset}
\end{figure}

In this section, we present additional statistics on the number of frames per class and the average proportion of each class within a sequence. 
Figure~\ref{fig_frame_num_over_class_num_of_gait3d_parsing_dataset} indicates that the majority of frames contain $7\sim10$ human body parts, which reflects the high quality of Gait3D~\cite{cvpr/gait3d_v1} dataset. 
Through Figure~\ref{fig_mean_proportion_in_one_sequence_over_classes_of_gait3d_parsing_dataset}, we can find that 
1) the head, torso, left-leg, and right-leg are the most consistently represented body parts, accounting for almost 100\% in a sequence, while dress has the smallest proportion at close to 0\%. 
2) the proportion of left-hand and right-hand are relatively low because hands are often occupied with various objects in real-world scenes and can be easily occluded due to their small area. 
3) There is a symmetry between left-arm/hand/leg/foot and right-arm/hand/leg/foot, which corresponds to the characteristic feature of humans walking alternately from left to right. 
Combining the above observations, we conclude that legs are more stable than hands in real-world scenes, indicating that more reliable dynamic information can be extracted from the legs. 

\section{Appendix: Exemplar Results}
In this section, we provide more detailed exemplar results of GaitBase~\cite{opengait} + Silhouette, GaitBase~\cite{opengait} + Parsing, and our ParsingGait. 
Specifically, we sample 24 consecutive frames from each sequence for visualization. 
The top row displays the query sequence with blue bounding boxes, while the subsequent rows show the top-8 gallery sequences ranked by their similarity to the query sequence. 
The correctly matched results are marked with green bounding boxes, while the incorrect ones are marked with red bounding boxes.

\begin{figure*}[t]
  \centering
   \includegraphics[width=1.0\linewidth]{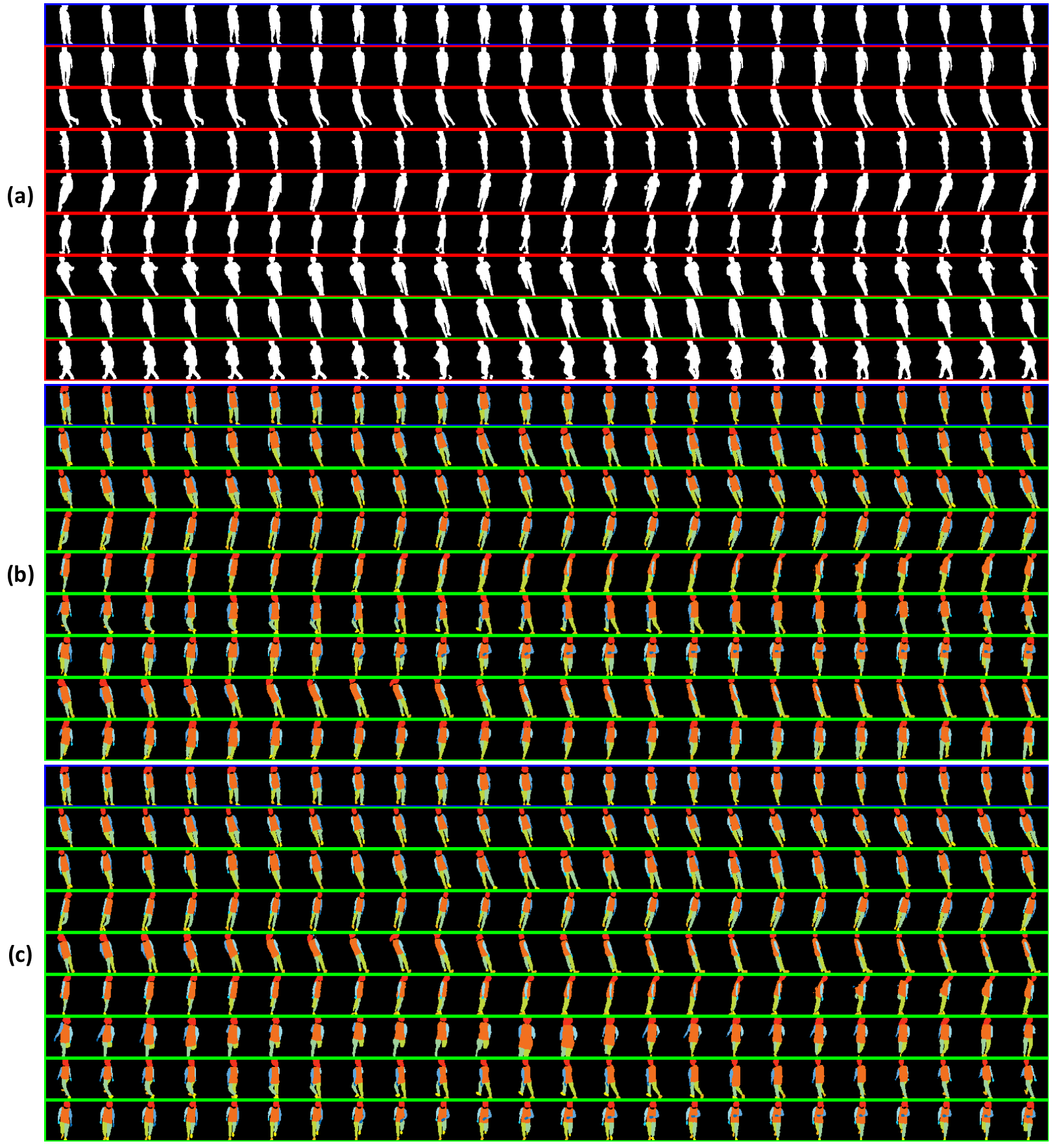}
   \caption{
   Exemplar results of (a) GaitBase+Silhouette, (b) GaitBase+Parsing, and (c) ParsingGait. 24 consecutive frames are sampled from each sequence for visualization. This case shows that the Gait Parsing Sequence (GPS) can help the model cope with situations of similar body shapes. (Best viewed in color.)
   }
   \label{fig_vis_exemplar_supp_1}
\end{figure*}

\begin{figure*}[t]
  \centering
   \includegraphics[width=1.0\linewidth]{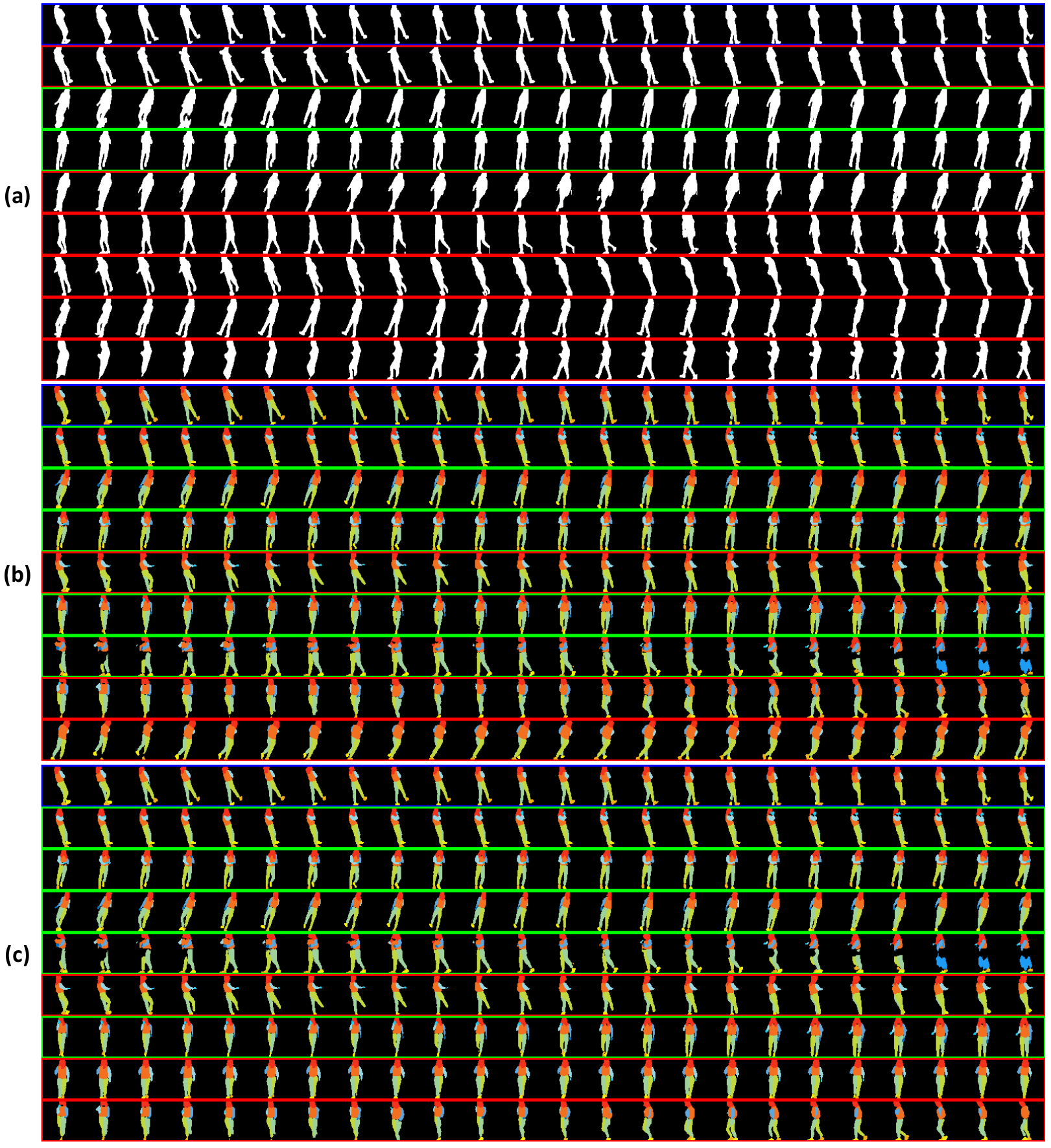}
   \caption{
   Exemplar results of (a) GaitBase+Silhouette, (b) GaitBase+Parsing, and (c) ParsingGait. 24 consecutive frames are sampled from each sequence for visualization. This case shows that the Gait Parsing Sequence (GPS) can help the model cope with situations of similar poses. (Best viewed in color.)
   }
   \label{fig_vis_exemplar_supp_2}
\end{figure*}

\begin{figure*}[t]
  \centering
   \includegraphics[width=1.0\linewidth]{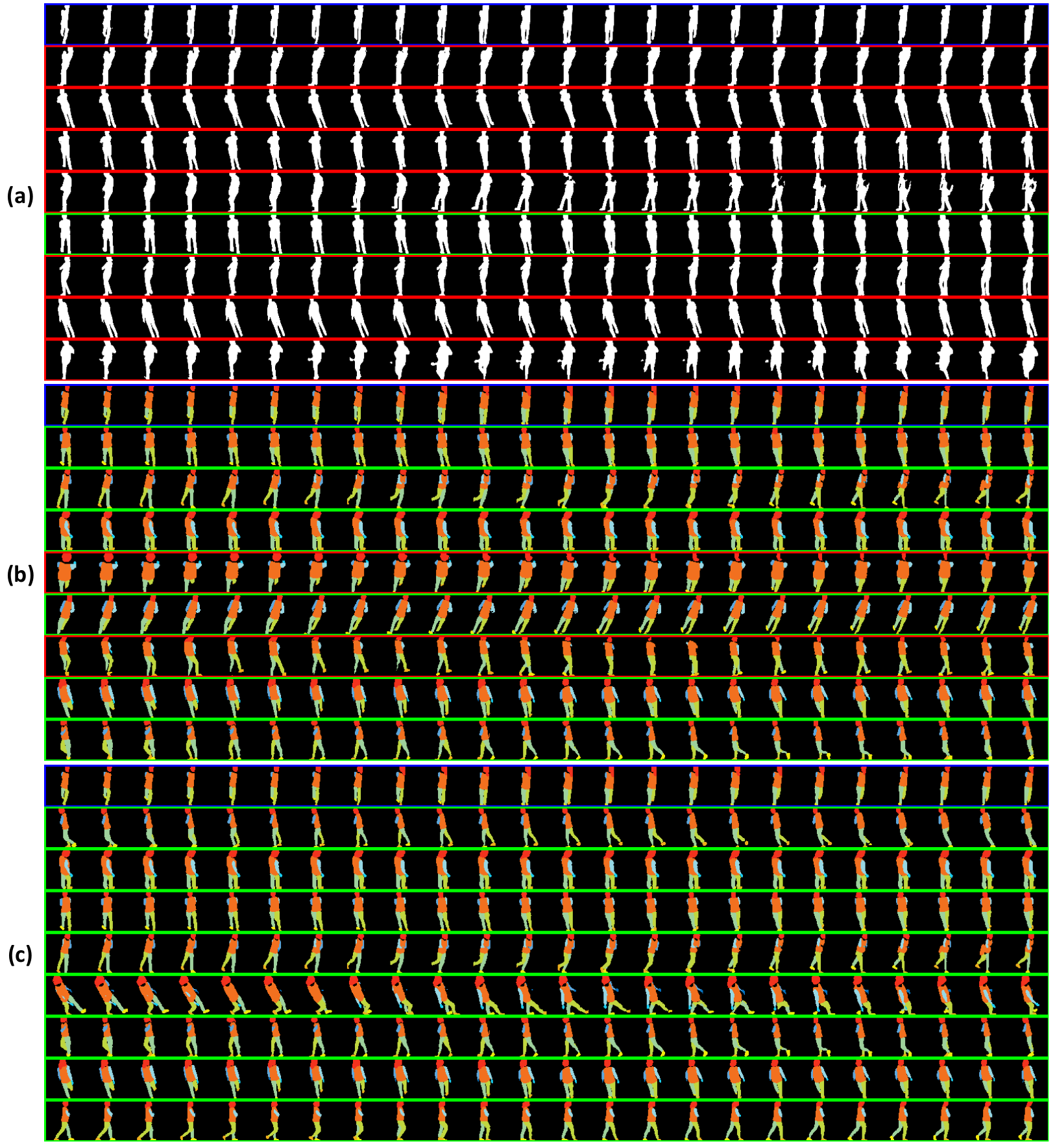}
   \caption{
   Exemplar results of (a) GaitBase+Silhouette, (b) GaitBase+Parsing, and (c) ParsingGait. 24 consecutive frames are sampled from each sequence for visualization. This case shows that the Gait Parsing Sequence (GPS) can help the model to alleviate the viewpoint problem. (Best viewed in color.)
   }
   \label{fig_vis_exemplar_supp_3}
\end{figure*}

\begin{figure*}[t]
  \centering
   \includegraphics[width=1.0\linewidth]{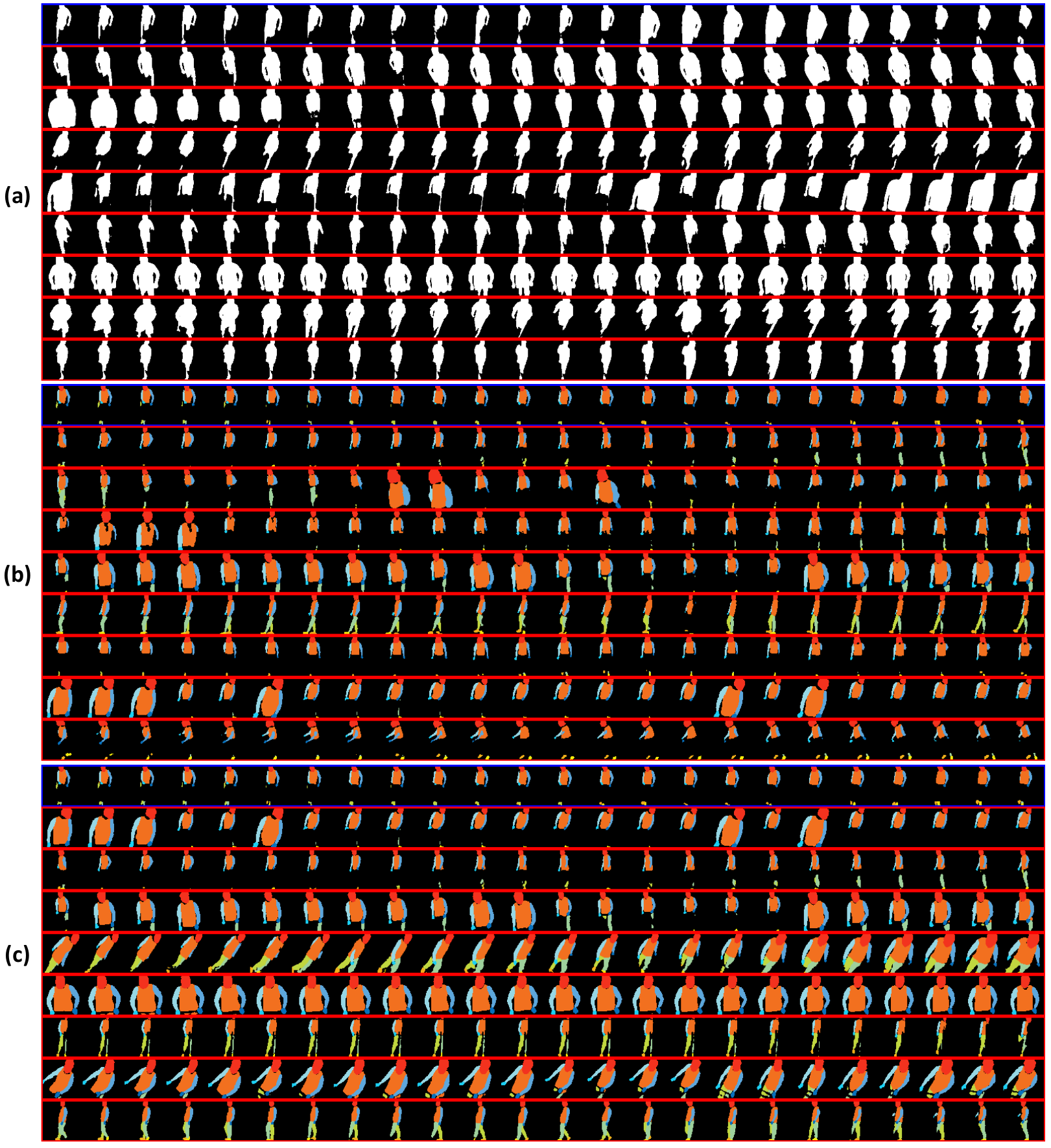}
   \caption{
   A bad case where individuals are seriously occluded. 
   It indicates that occlusion is a very challenging condition of gait recognition in the wild. 
   (Best viewed in color.)
   }
   \label{fig_vis_exemplar_supp_4}
\end{figure*}

From Figure~\ref{fig_vis_exemplar_supp_1} to~\ref{fig_vis_exemplar_supp_3}, we find that Gait Parsing Sequence (GPS) with high information entropy can assist the model in learning useful gait features in the case with similar body shape, similar pose, and cross-viewpoint, thereby improving the recognition accuracy. 
Our ParsingGait achieves the highest performance among the compared methods. 
However, Figure~\ref{fig_vis_exemplar_supp_4} illustrates a bad case where individuals are severely occluded, highlighting that occlusion remains a significant challenge in gait recognition in the wild. 
In the matching results shown in Figure~\ref{fig_vis_exemplar_supp_4}, we observe some frame loss, which could be due to detection omission resulting from severe occlusion in real-world scenes. 
Such frame loss could potentially affect the accuracy of gait recognition in the wild.

\section{Appendix: Future Work}

Despite the proposed ParsingGait method for gait recognition in the wild, there are still many potential directions for this challenging task using Gait Parsing Sequence (GPS): 
\begin{itemize}
    \item One direction is to explore the design of a deep CNN or transformer-based model to learn more discriminative features directly from GPS because GPS provides higher entropy information than the silhouette-based sequence. 
    \item Second direction is how to deal with the occlusion problem in a real-world scene. 
    A feasible solution would be using unoccluded frames to predict the occluded ones. 
    \item Another interesting direction is to develop a more efficient method to correlate parsed human parts information. 
    \item As discussed in Section~\ref{more_statistics_of_gait3d_parsing}, there is also a need to investigate how to robustly model dynamic gait features from human body parts like legs in real-world scenes. 
\end{itemize}

\end{document}